\documentclass{article}

\usepackage[final]{neurips_2019}

\usepackage[utf8]{inputenc} %
\usepackage[T1]{fontenc}    %
\usepackage{hyperref}       %
\usepackage{url}            %
\usepackage{booktabs}       %
\usepackage{amsmath}        %
\usepackage{amsfonts}       %
\usepackage{nicefrac}       %
\usepackage{microtype}      %
\usepackage{mathtools}      %
\usepackage{siunitx}        %
\usepackage[capitalize,nameinlink,noabbrev]{cleveref}
\usepackage{xspace}       %
\usepackage{natbib}
\usepackage{multicol}
\usepackage[utf8]{inputenc}%
\usepackage{tabularx} 
\usepackage{mathtools, nccmath}
\usepackage{tikz}
\usepackage{pdfpages}
\usepackage{algorithm}
\usepackage{algpseudocode}
\usepackage{multirow}
\usepackage{mathabx}
\usepackage{pifont}
\usepackage{bm}
\usepackage{pgfplots}
\usepackage{graphicx}
\usepackage{xcolor}
\usepackage{wrapfig}
\usepackage{lipsum}
\usepackage{listings}
\usepackage{xr}
\usepackage{xr-hyper}
\usepackage{hyperref}
\usepackage[all]{hypcap}

\makeatletter
\newcommand{\leqnomode}{\tagsleft@true\let\veqno\@@leqno}%
\newcommand{\reqnomode}{\tagsleft@false\let\veqno\@@eqno}%
\newcommand*{\compress}{\@minipagetrue}
\makeatother

\title{Learning Dynamics of Attention:\\ Human Prior for Interpretable Machine Reasoning}

\author{%
  Wonjae Kim \\
  Kakao Corporation \\
  Pangyo, Republic of Korea \\
  \texttt{dandelin.kim@kakaocorp.com} \\
  \And
  Yoonho Lee \\
  Kakao Corporation \\
  Pangyo, Republic of Korea \\
  \texttt{eddy.l@kakaocorp.com} \\
}

\usepackage{color}
\definecolor{orange}{rgb}{1,0.3,0}
\definecolor{copper}{rgb}{1,.62,.40}
\definecolor{hotpink}{rgb}{1,0,0.5}
\definecolor{darkgreen1}{rgb}{0, .35, 0}
\definecolor{darkgreen}{rgb}{0, .6, 0}
\definecolor{darkred}{rgb}{.75,0,0}

\newcommand{\be}{\begin{eqnarray}}
\newcommand{\ee}{\end{eqnarray}}
\newcommand{\bee}{\begin{eqnarray*}}
\newcommand{\eee}{\end{eqnarray*}}

\newcommand{\matrixb}{\left[ \begin{array}}
\newcommand{\matrixe}{\end{array} \right]}

\DeclarePairedDelimiterX{\infdivx}[2]{(}{)}{%
  #1\;\delimsize\|\;#2%
}

\newcommand{\aff}[2]{\mathbf{W}^{{#1} \times {#2}}}

\newcommand{\NewCross}[1]{
\begin{tikzpicture}[#1,line width=.19931109ex,inner sep=0pt]
\draw (0ex,0ex) -- (1ex,1ex);
\draw (0ex,1ex) -- (1ex,0ex);
\draw node[inner sep=0,text opacity=0] at (0.5ex,0.5ex) {X};
\end{tikzpicture}
}

\newcommand{\NewCircle}[1]{
\begin{tikzpicture}[#1,line width=.19931109ex,inner sep=0pt]
\draw (0.5ex, 0.5ex) circle (0.5ex);
\draw node[inner sep=0,text opacity=0] at (0.5ex,0.5ex) {O} ;
\end{tikzpicture}
}

\usetikzlibrary{arrows}
\usetikzlibrary{positioning}
\usetikzlibrary{calc}
\usetikzlibrary{fit}
\usetikzlibrary{automata}
\usetikzlibrary{shapes.geometric}
\usetikzlibrary{arrows.meta}
\usetikzlibrary{backgrounds}
 
\definecolor{amethyst}{rgb}{0.6, 0.4, 0.8}
\definecolor{darkmidnightblue}{rgb}{0.0, 0.2, 0.4}
\definecolor{darkpastelred}{rgb}{0.76, 0.23, 0.13}
\definecolor{deepblue}{rgb}{0,0,0.5}
\definecolor{deepred}{rgb}{0.6,0,0}
\definecolor{deepgreen}{rgb}{0,0.5,0}

\usepackage{hyperref}
\hypersetup{
    colorlinks=true,
    urlcolor=amethyst,
    linkcolor=darkpastelred,
    citecolor=darkmidnightblue
}

\DeclareFixedFont{\ttb}{T1}{txtt}{bx}{n}{7} %
\DeclareFixedFont{\ttm}{T1}{txtt}{m}{n}{7}  %

\newcommand\pythonstyle{\lstset{
language=Python,
basicstyle=\ttm,
otherkeywords={self},             %
keywordstyle=\ttb\color{deepblue},
emph={MyClass,__init__},          %
emphstyle=\ttb\color{deepred},    %
stringstyle=\color{deepgreen},
frame=tb,                         %
showstringspaces=false            %
}}

\lstnewenvironment{python}[1][]
{
\pythonstyle
\lstset{#1}
}
{}

\newcommand*{\tikzmk}[1]{\tikz[remember picture,overlay,] \node (#1) {};\ignorespaces}
\newcommand{\boxit}[1]{\tikz[remember picture,overlay]{\node[yshift=2.2pt,xshift=-7.2pt,fill=#1,opacity=.25,fit={(A)(B)}] {};}\ignorespaces}
\colorlet{pink}{red!40}
\colorlet{blue}{cyan!60}

\newtheorem{definition}{Definition}

\begin{document}
\maketitle

\begin{abstract}
Without relevant human priors, neural networks may learn uninterpretable features.
We propose \textbf{D}ynamics of \textbf{A}ttention for \textbf{F}ocus \textbf{T}ransition (DAFT) as a human prior for machine reasoning.
DAFT is a novel method that regularizes attention-based reasoning by modelling it as a continuous dynamical system using neural ordinary differential equations.
As a proof of concept, we augment a state-of-the-art visual reasoning model with DAFT.
Our experiments reveal that applying DAFT yields similar performance to the original model while using fewer reasoning steps, showing that it implicitly learns to skip unnecessary steps.
We also propose a new metric, \textbf{T}otal \textbf{L}ength of \textbf{T}ransition (TLT), which represents the effective reasoning step size by quantifying how much a given model's focus drifts while reasoning about a question.
We show that adding DAFT results in lower TLT, demonstrating that our method indeed obeys the human prior towards shorter reasoning paths in addition to producing more interpretable attention maps.
Our code is available at \url{https://github.com/kakao/DAFT}.
\end{abstract}

\section{Introduction} \label{sec:intro} \begin{wrapfigure}{R}{0.345\textwidth}
    \vspace{-0.4cm}
    \centering
    \includegraphics[width=\linewidth]{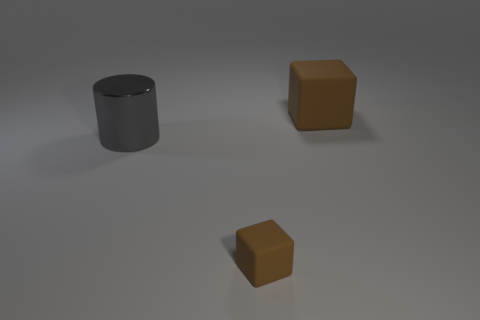}
    \vspace{-0.55cm}
    \caption{
      "What color is the cube nearest to the cylinder?" can be answered without knowing the relative location of objects.
    }
    \label{fig:clevr_example}
    \vspace{-0.5cm}
\end{wrapfigure}

We focus on the task of visual question answering (VQA) \citep{agrawal2015vqa},
which tests visual reasoning capability by measuring how well a model can answer a question by composing supporting facts from a given image.
An example of such a question-image pair from the CLEVR dataset \citep{johnson2017clevr} is shown in \cref{fig:clevr_example}.
One strategy for solving this example is to first find the cube that the question is referring to, and then reporting its color.
However, the first step would be unnecessary since all cubes in the image are brown.
Questions with such redundancy can be pruned using the complete scene graph. 
While complete scene graphs are provided in CLEVR, this process is not applicable to real-world images since obtaining their scene graphs is notoriously hard.

The motivation behind training visual reasoning models on the VQA task is to obtain a model that reasons about images similarly to humans.
We prefer human-like reasoning because such reasoning is believed to be concise and effective.
Conversely, we can say that a model's reasoning is ineffective if it retains and references facts that are irrelevant to the given question, even if its answers are correct.
This work is motivated by the question: "How can we measure the degree to which a given model only uses necessary information?"

To this end, we adopt the minimum description length (MDL) principle \citep{rissanen1978modeling}, which formalizes Occam's razor and is also a relaxation of Kolmogorov complexity \citep{kolmogorov1963tables}.
This principle states that the best hypothesis for a given data is the one that provides the shortest description of it.
The MDL framework offers two benefits: 
(1) it encourages models to more tightly compress the data, and 
(2) incentivizes more interpretable models.
The first claim comes naturally from the definition of MDL principle since minimum description length is the optimal compression of the data.
The inverse relation between interpretability and compression has been demonstrated empirically by numerous works in cognitive neuroscience starting from the work of \cite{hochberg1953quantitative} to its modern follow-up studies \citep{feldman2009bayes, feldman2016simplicity}.

We thus aim for a VQA method which produces solutions with short description length (in the context of VQA, we also call this a \textit{program}).
With the ground-truth program supervision, we can train a model that produces short and effective programs.
We aim for an end-to-end learnable reasoning model which produces solutions with short description length.
In VQA, such solutions can be seen as learned versions of \textit{explicit} programs, for which we have ground-truth supervision on synthetic datasets such as CLEVR.
However such programs are nontrivial to obtain for non-relational questions and can be ill-posed for images with incomplete scene graph.
Instead of using the ground-truth program as supervision, we construct a model that \textit{continuously} changes its attention over time, which we experimentally show shifts focus less compared to previous models.
This is motivated by experiments \citep{vendetti2014evolutionary} which show that the focus (i.e. attention) of the lateral frontoparietal network on the context changes continuously.
Our model, \textbf{Dynamics of Attention for Focus Transition} (DAFT), models the infinitesimal change of its attention at each timepoint.
Since the resulting attention map is differentiable, it is a continuous funtion over time.

The solution of the initial value problem (IVP) specified by DAFT is a continuous function which specifies the attention map of the model at each point in time.
Note that such IVP solutions can be used as a drop-in replacement for any of the discrete attention mechanisms used by previous machine reasoning models.
While DAFT is applicable to any attention-based step-wise reasoning model, we applied it to the MAC network \citep{hudson2018compositional}, a state-of-the-art visual reasoning model, to show how this human prior acts in a holistic model.
In addition to DAFT, we propose \textbf{Total Length of Transition} (TLT), a metric that quantifies the description length of a given attention map, thus measuring the degree to which a model follows the MDL principle.
TLT enables a direct quantitative comparison between the quality of reasoning of different models,
unlike previous works which only inspected the reasoning of VQA models qualitatively by visualizing attention maps.

This paper is organized as follows.
We describe background concepts and their connections to our work in \cref{sec:background}.
We propose DAFT with a detailed explanation of how to adapt DAFT to existing models in \cref{sec:daft}.
We present experiments in \cref{sec:experiments}, and importantly, we define and measure TLT in \cref{subsec:measure}.
We conclude the paper with future directions in \cref{sec:conclusion}.

\section{Background} \label{sec:background} Our work encompasses multiple disciplines of machine learning including visual question answering, interpretable machine learning, and neural ordinary differential equations.
In this section, we summarize each and explain how they are related to our work.

\subsection{Visual Question Answering}
Machine reasoning tasks were proposed to test whether algorithms can demonstrate high-level reasoning capabilities once believed to be only possible for humans \citep{bottou2014machine}.
Given knowledge base $\mathbf{K}$ and task description $\mathbf{Q}$, the model composes supporting facts from $\mathbf{K}$ to accomplish the task described by $\mathbf{Q}$.
Visual question answering (VQA) is an instance of a machine reasoning task in the visual domain where $\mathbf{K}$ is an image and $\mathbf{Q}$ is a question about the image ($\mathbf{K}$).

Approaches for solving VQA vary widely on which supervisory signals are given.
The usual supervisory signals in VQA comprise images, questions, answers, programs, and object masks.
Following \cite{mao2018neuro}, we denote program and object mask supervisions as \textit{additional supervision} and others as \textit{natural supervision}.
Natural supervision signals are only signals that all VQA datasets have in common \citep{agrawal2015vqa, krishna2017visual, goyal2017making, hudson2019gqa}, because the additional supervisions are generally hard to acquire.

Given additional supervision, the VQA model can \textit{infer and execute its program on the given scene graph} (i.e. \textit{symbolic} models) \citep{johnson2017inferring}.
We refer the reader to \cref{app:related} for further exposition on models that take this approach.
Although symbolic models often employ a neural attention mechanism for program execution (e.g., module networks \citep{andreas2016neural, hu2017learning,johnson2017inferring, mascharka2018transparency}), such attention is not necessary if the perfect scene graph can be inferred \citep{yi2018neural}.

On the other hand, non-symbolic models, which only use natural supervision, generally all employ some form of attention onto the features of $\mathbf{K}$ from the features of $\mathbf{Q}$ \citep{xiong2016dynamic,hudson2018compositional}.
Although non-symbolic attention-based models achieve competitive state-of-the-art performance in VQA datasets without additional supervisions (\cref{table:baselines}), no discussions on the effectiveness of its latent program have been made so far.
Our work investigates this question by quantitatively measuring the quality of these latent programs and proposes a model that improves on this measure, similarly to how symbolic models are optimized for the effectiveness of their programs.

\subsection{Human Prior and Interpretability}
With the growing demands for interpretable machine learning, attention-based models demonstrated their interpretability by showing their attention map visualizations.
However, \cite{ilyas2019adversarial} claimed that without a human prior, neural networks eventually learn \textit{useful but non-robust features} which are highly predictive for the model but not useful for humans.
Concurrently, \cite{poursabzi2018manipulating} and \cite{lage2018human} empirically show how human prior affects the interpretability of the model.

More concretely in VQA, the length of description has no meaning for the model as long as it gets the right answer.
For example, \citep{hudson2018compositional} observed that increasing reasoning step length leaves the model's performance intact (\textit{useful}) but their attention maps became uninterpretable (\textit{non-robust}).
To solve this problem, we propose DAFT in \cref{sec:daft} to embed the human reasoning prior of continuous focus transition in attention-based machine reasoning models.

Another problem is that there exists no method to quantitatively measure the interpretability of attention-based models.
This is because interpretability is fundamentally qualitative, and by principle, it can only be measured via a user study.
However, user studies cannot scale to large datasets such as CLEVR \citep{johnson2017clevr} GQA \citep{hudson2019gqa}.

Thus we propose TLT as a quantitative and scalable proxy for interpretability, backed with empirical evidence \citep{hochberg1953quantitative, feldman2009bayes, feldman2016simplicity} in \cref{subsec:measure}.

\subsection{Neural Ordinary Differential Equations} \label{subsec:ode}
Recent work on residual networks \citep{lu2017beyond, haber2017stable, ruthotto2018deep} interpret residual connections as an Euler discretization of a continuous transformation through time.
Motivated by this interpretation, \cite{chen2018neural} generalized residual networks by using more sophisticated black-box ODE solvers such as \texttt{dopri5} \citep{dormand1980family} and proposed a new family of neural networks called neural ordinary differential equations (neural ODEs).

Adaptive-step ODE solvers such as \texttt{dopri5} perform multiple function evaluations to adapt their step size,
shortening the steps when the gaps between estimations increase and lengthening otherwise.
One can find resemblance between adaptive-step ODE solvers and adaptive computation time methods used in recurrent networks \citep{graves2016adaptive, dehghani2018universal}.
However, as mentioned in \citep{chen2018neural}, adaptive-step ODE solvers offer more well-studied, computationally cheap and generalizable rules for adapting the amount of computation.
We applied neural ODEs to modeling the infinitesimal change of the model's attention.

\cite{dupont2019augmented} stated that the homeomorphism of neural ODEs greatly restricts the representation power of the dynamics and show a number of functions which cannot be represented by the family of neural ODEs.
They showed that by augmenting the feature space by adding empty dimensions, the dynamics of neural ODEs can be simplified.
To show its efficacy, they measured the number of function evaluation (NFE) during training, since complex dynamics requires exponentially many function evaluations while solving IVP.
They showed that augmented neural ODEs yield a gradually growing NFE during training while their non-augmented counterpart has an NFE that grows exponentially.
We show the connection between our model (DAFT) and augmented neural ODEs in \cref{sec:daft}.

\section{Dynamics of Attention for Focus Transition} \label{sec:daft} \begin{center}
\begin{algorithm}[ht]
    \caption{Memory Update Procedure of MAC}
    \label{alg:mac}
    \begin{algorithmic}[1]
        \Require current time $t_0$, next time $t_1$, current memory $\mathbf{m}_{t_0}$, contextualized question $\mathbf{cw} \in \mathbb{R}^{L \times d}$, atomic question $\mathbf{q} = [\overleftarrow{\mathbf{cw}_1}, \overrightarrow{\mathbf{cw}_L}]$, knowledge base $\mathbf{K} \in \mathbb{R}^{S \times d}$ \label{alg:daft:input}
        \Ensure next memory $\mathbf{m}_{t_1}$\label{alg:daft:output}
        \State \tikzmk{A} $\mathbf{a}_{t_1} = \aff{1}{d} (\aff{d}{d}_{t_1} \mathbf{q} \odot \mathbf{cw})$ \Comment{get \textit{attention logit} on $\mathbf{cw}$} \tikzmk{B} \label{alg:mac:attention}
        \boxit{violet!40}
        \State $\mathbf{c}_{t_1} = \sum_{i=0}^{L} \text{softmax}(\mathbf{a}_{t_1})(i) \odot \mathbf{cw}(i)$ \Comment{get \textit{control} vector} \label{alg:mac:control}
        \State $\mathbf{rq}_{t_1} = \aff{1}{d} (\aff{d}{2d} [\aff{d}{d} \mathbf{K} \odot \aff{d}{d} \mathbf{m}_{t_1}, \mathbf{K}] \odot \mathbf{c}_{t_1})$ \Comment{get \textit{attention logit} on $\mathbf{K}$} \label{alg:mac:read1}
        \State $\mathbf{r}_{t_1} = \sum_{i=0}^{S} \text{softmax}(\mathbf{rq}_{t_1})(i) \odot \mathbf{K}(i)$ \Comment{get \textit{information} vector} \label{alg:mac:read2}
        \State $\mathbf{m}_{t_1} = \aff{d}{2d}[\mathbf{r}_{t_1}, \mathbf{m}_{t_0}]$ \Comment{get \textit{memory} vector} \label{alg:mac:write}
    \end{algorithmic}
\end{algorithm}
\end{center}

\paragraph{The MAC Network}
We briefly review the MAC network \citep{hudson2018compositional}.
It consists of three subunits (control, read, and write) which rely on each other to perform visual reasoning.
\cref{alg:mac} describes how the MAC network updates its memory vector given its inputs.
Given initial memory vector $\mathbf{m}_0$, it performs a fixed number ($T$) of iterative memory updates to produce the final memory vector $\mathbf{m}_T$.
MAC infers answer logits by processing the concatenation of $\mathbf{q}$ and $\mathbf{m}_T$ through a 2-layer classifier : $\aff{1}{d}(\aff{d}{2d}[\mathbf{q},\mathbf{m}_T])$\footnote{
We omit biases and nonlinearities for brevity.
}.
The original work optionally considers additional structures inside the write unit.
Unlike the description in the original paper, previous control $\mathbf{c}_{t-1}$ is not used when computing the current control $\mathbf{c}_t$ in the official impelementation\footnote{\url{https://github.com/stanfordnlp/mac-network/blob/master/configs/args.txt}}.
Please refer the original paper \citep{hudson2018compositional} for the details.

\begin{center}
\begin{algorithm}[ht]
    \caption{Memory Update Procedure of DAFT MAC}
    \label{alg:daft}
    \begin{algorithmic}[1]
        \Require current time $t_0$, next time $t_1$, current memory $\mathbf{m}_{t_0}$, contextualized question $\mathbf{cw} \in \mathbb{R}^{L \times d}$, atomic question $\mathbf{q} = [\overleftarrow{\mathbf{cw}_1}, \overrightarrow{\mathbf{cw}_L}]$, knowledge base $\mathbf{K} \in \mathbb{R}^{S \times d}$, current attention logit $\mathbf{a}_{t_0}$ \label{alg:daft:input}
        \Ensure next memory $\mathbf{m}_{t_1}$, next attention logit $\mathbf{a}_{t_1}$ \label{alg:daft:output}
        \State \tikzmk{A} \textbf{def} f($\mathbf{a}_t$, $t$): \Comment{\textbf{Define DAFT}} \label{alg:daft:dynamic1}
        \State \hskip1.5em \textbf{return} $\aff{1}{(d+1)}[\aff{d}{(d+1)}[t, \mathbf{q}] \odot \mathbf{cw}, \mathbf{a}_t]$ \Comment{compute $\frac{d\mathbf{a}_t}{dt}$} \label{alg:daft:dynamic2}
        \State $\mathbf{a}_{t_1} = \mathbf{a}_{t_0} + \int_{t_0}^{t_1} f(\mathbf{a}_{t}, t) dt = \text{ODESolve}(\mathbf{a}_t, f, t_0, t_1)$  \Comment{Solve IVP using DAFT} \tikzmk{B} \label{alg:daft:ivp}
        \boxit{violet!40}
        \State $\mathbf{c}_{t_1} = \sum_{i=0}^{L} \text{softmax}(\mathbf{a}_{t_1})(i) \odot \mathbf{cw}(i)$ \label{alg:daft:control}
        \State $\mathbf{rq}_{t_1} = \aff{1}{d} (\aff{d}{2d} [\aff{d}{d} \mathbf{K} \odot \aff{d}{d} \mathbf{m}_{t_0}, \mathbf{K}] \odot \mathbf{c}_{t_1})$ \label{alg:daft:read_query}
        \State $\mathbf{r}_{t_1} = \sum_{i=0}^{S} \text{softmax}(\mathbf{rq}_{t_1})(i) \odot \mathbf{K}(i)$ \label{alg:daft:read}
        \State $\mathbf{m}_{t_1} = \aff{d}{2d}[\mathbf{r}_{t_1}, \mathbf{m}_{t_0}]$ \label{alg:daft:write}
    \end{algorithmic}
\end{algorithm}
\end{center}

\paragraph{The DAFT MAC Network}
We now introduce Dynamics of Attention for Focus Transition (DAFT) and its application to MAC;
we call this augmented MAC model as DAFT MAC.

\cref{alg:daft} shows the memory update procedure of DAFT MAC and the definition of DAFT in full detail.
We colored the differences in \cref{alg:mac} and \cref{alg:daft}.
We point out that DAFT can just as easily be applied to any other memory-augmented model by replacing discrete attention with a neural ODE as we have done in \cref{alg:daft}.

Unlike MAC, the memory update procedure of DAFT MAC requires the previous attention logit, meaning we need to define the initial attention logit.
We use a zero vector as the initial attention logit $\mathbf{a}_0$ to produce uniformly distributed attention weight, assuming the model's focus distributed evenly at the start of reasoning.

\begin{figure}[ht]
\centering
\input{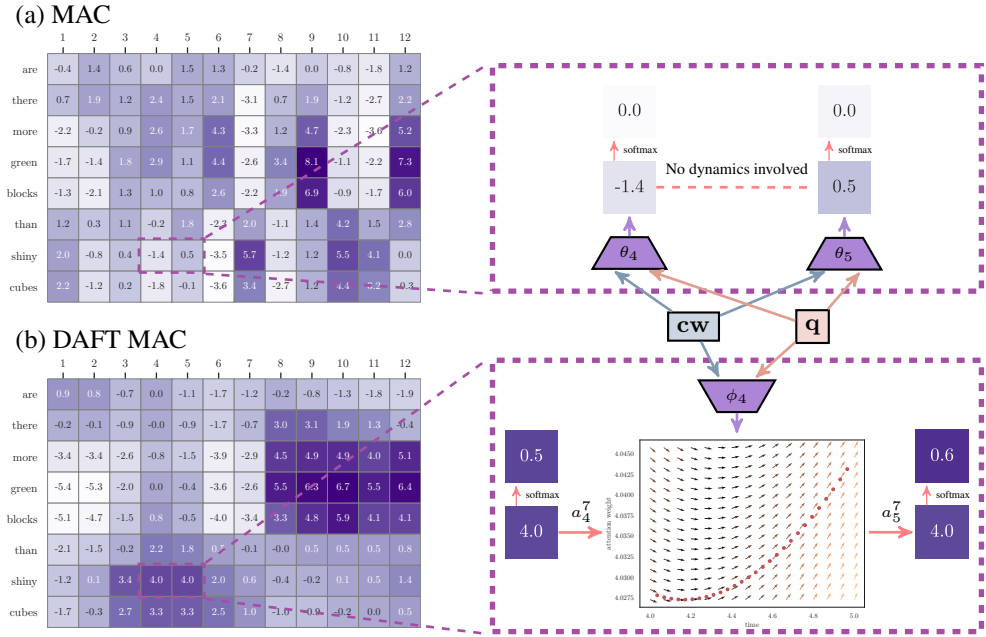}
\caption{
  A graphical description of how attention logits change in MAC and DAFT MAC for an example in the CLEVR dataset.
  The question is \textit{"are there more green blocks than shiny cubes?"}.
  Attention logits maps of 12-step (a) MAC and (b) DAFT MAC are shown.
  The right side shows a magnified view of a single step of attention shift on the word \textit{shiny}.
}
\label{fig:daft}
\end{figure}

\cref{fig:daft} shows the difference between MAC and DAFT MAC graphically.
While MAC has no explicit connection between adjacent logits, DAFT MAC computes the next attention logit by solving the IVP starting from the current attention logit.
Note that the actual attention weight is the softmax-ed value of attention logits.
Since softmax computes the size of a logit relative to other logits, small changes in attention logit can result in a large difference in the attention weight (See \cref{fig:interpretation} for 
a visualization of the attention weight).

\paragraph{Connection to Augmented Neural ODEs} As shown in \cref{fig:daft}, every token $\mathbf{cw}$ and its question $\mathbf{q}$ acts as a condition on the dynamics.
Empirically, we found that the conditionally generated ODE dynamics do not suffer from number of function evaluations (NFE) explosion while solving IVP until the end of training (see \cref{fig:nfe} in the appendix for more details on NFE).
This is remarkable since the VQA is incomparably more complex than the toy problems treated in previous works.
We thus argue that these conditional ODE dynamics are another form of augmentation for neural ODEs as it differs from the previous unconditioned neural ODEs \citep{chen2018neural, dupont2019augmented}.

\paragraph{Alternative Ways to Restrict Focus Transition}
Besides DAFT, we tested two simple alternatives to restrict the model's transition of attention.
The first is to introduce a residual connection at each attention step,
which is equivalent to DAFT using a single-step Euler solver during training.
We observed significant drops in accuracy, 
and attention maps of this model deffered all transitions to the last few steps.
We attribute this phenomenon to this residual model having insufficient expressive power compared to the complex visual information being incorporated at each step.
Our second baseline is to add the TLT itself to objective function with Lagrange multiplier $\lambda$.
This model significantly harmed performance for every $\lambda$ in the wide range we tested.

\section{Experiments} \label{sec:experiments} We conducted our experiments on the CLEVR\footnote{\url{https://cs.stanford.edu/people/jcjohns/clevr/}} \citep{johnson2017clevr} and GQA\footnote{\url{https://cs.stanford.edu/people/dorarad/gqa/about.html}} \citep{hudson2019gqa} datasets.
For brevity we put the results from GQA dataset in the \cref{app:gqa}.

To evaluate the efficacy of DAFT, we conducted experiments on two different criteria: performance (accuracy and run-time) and interpretability.
For a fair comparison, we used the same hyperparameters as the original MAC network \citep{hudson2018compositional} and closely followed their experimental setup.
The only difference from the original MAC network is in the computation of attention logits and control vectors (highlighted in purple in \cref{alg:daft}).
We list implementation details in \cref{app:impl}.

\subsection{CLEVR Dataset} \label{subsec:dataset} \begin{table}[ht]
    \centering
    \fontsize{8.8pt}{8.8pt}\selectfont
    \caption{
        Accuracies on the CLEVR dataset of baselines with various additional annotation types ($\textbf{P}$ for program and $\textbf{M}$ for object mask annotation) and our model.
        $D$ denotes depth of the inferred program.
        $\bm{\triangle}$ means that additional annotation is implicitly provided through the pretrained object detector such as Mask R-CNN.
        }
    \begin{tabular}{lccccccccc}
    \toprule

    \multirow{2}{*}{Model}
    & \multicolumn{2}{c}{Anno.}
    & \multirow{2}{*}{\shortstack{\#\\Step}}
    & \multirow{2}{*}{Avg.}
    & \multirow{2}{*}{Count}
    & \multirow{2}{*}{Exist}
    & \multirow{2}{*}{\shortstack{Cmp.\\Num.}}
    & \multirow{2}{*}{\shortstack{Query\\Attr.}}
    & \multirow{2}{*}{\shortstack{Cmp.\\Attr.}} \\

    & \multicolumn{1}{c}{\textbf{P}} & \multicolumn{1}{c}{\textbf{M}}
    &
    &
    &
    &
    &
    & 
    & \\
    \midrule
    Human \citep{johnson2017clevr} & -- & -- & -- & 92.6 & 86.7 & 96.6 & 86.5 & 95.0 & 96.0 \\
    \midrule
    NMN \citep{andreas2016neural} & $\NewCircle{scale=1.4}$ & $\NewCross{scale=1.4}$ & $D$ & 72.1 & 52.5 & 79.3 & 72.7 & 79.0 & 78.0 \\
    N2NMN \citep{hu2017learning} & $\NewCircle{scale=1.4}$ & $\NewCross{scale=1.4}$ & $D$ & 88.8 & 68.5 & 85.7 & 84.9 & 90.0 & 88.8 \\
    IEP \citep{johnson2017inferring} & $\NewCircle{scale=1.4}$ & $\NewCross{scale=1.4}$ & $D$ & 96.9 & 92.7 & 97.1 & 98.7 & 98.1 & 98.9 \\
    DDRprog \citep{suarez2018ddrprog} & $\NewCircle{scale=1.4}$ & $\NewCross{scale=1.4}$ & $D$ & 98.3 & 96.5 & 98.8 & 98.4 & 99.1 & 99.0 \\
    TbD \citep{mascharka2018transparency} & $\NewCircle{scale=1.4}$ & $\NewCross{scale=1.4}$ & $D$ & 99.1 & 97.6 & 99.2 & 99.4 & 99.5 & 99.6 \\
    \midrule
    NS-VQA \citep{yi2018neural} & $\NewCircle{scale=1.4}$ & $\NewCircle{scale=1.4}$ & $D$ & 99.8 & 99.7 & 99.9 & 99.9 & 99.8 & 99.8 \\
    \midrule
    NS-CL \citep{mao2018neuro} & $\NewCross{scale=1.4}$ & $\bm{\triangle}$ & $D$ & 98.9 & 98.2 & 99.0 & 98.8 & 99.3 & 99.1 \\
    \midrule
    RN \citep{santoro2017simple} & $\NewCross{scale=1.4}$ & $\NewCross{scale=1.4}$ & 1 & 95.5 & 90.1 & 97.8 & 93.6 & 97.1 & 97.9 \\
    FiLM \citep{perez2018film} & $\NewCross{scale=1.4}$ & $\NewCross{scale=1.4}$ & 4 & 97.6 & 94.5 & 99.2 & 93.8 & 99.2 & 99.0 \\
    MAC \citep{hudson2018compositional} & $\NewCross{scale=1.4}$ & $\NewCross{scale=1.4}$ & 12 & 98.9 & 97.2 & 99.5 & 99.4 & 99.3 & 99.5 \\
    \midrule
    DAFT MAC (Ours) & $\NewCross{scale=1.4}$ & $\NewCross{scale=1.4}$ & \textbf{4} & 98.9 & 97.2 & 99.5 & 98.3 & 99.6 & 99.3 \\
    \bottomrule
\end{tabular}
    \label{table:baselines}
\end{table}

CLEVR dataset was proposed to evaluate the visual reasoning capabilities of a model.
CLEVR includes five supervisory signals: images, questions, answers, programs, and object masks (in addition to ground-truth scene graphs).
Images in CLEVR are synthetic scenes containing objects with various attributes: size, material, color, shape.
Each image has multiple questions with corresponding answers to test relational and non-relational visual reasoning abilities.

We provide a survey of previous models for CLEVR in \cref{table:baselines}, showing the accuracy by question type in addition to what additional supervision is given to the model.
In total, CLEVR has 700K questions for training and 150K questions for validation and test split.
All accuracies and TLT measured in the following sections were evaluated on the 150K validation set.
\subsection{Performance} \label{subsec:performance} \begin{wrapfigure}{R}{0.5\textwidth}
  \centering
  \begin{tikzpicture}
    \begin{axis}[
        height=5cm,
        width=7.5cm,
        /pgf/bar width=9pt,%
        xtick={1,...,10},
        xticklabels={%
            2-step,
            3-step,
            4-step,
            5-step,
            6-step,
            8-step,
            },
        ylabel={accuracy (val)},
        y label style={at={(0.08,0.5)}},
        grid=major,
        ybar,
        legend pos=south east,
        extra y ticks=98.9,
        extra y tick labels={\tiny 98.9},
        extra y tick style={
            ymajorgrids=true,
            ytick style={
                /pgfplots/major tick length=0pt,
            },
            grid style={
                purple!80,
                dashed,
                /pgfplots/on layer=axis foreground,
            },
        },
        ]
    
    \addplot[
        fill=orange!50,
        draw=black,
        point meta=y,
        every node near coord/.style={inner ysep=5pt},
        error bars/.cd,
            y dir=both,
            y explicit
    ] 
    table [y error=error] {
        x y error
        1 94.6 1.68
        2 97.4 0.55
        3 98.6 0.42
        4 98.7 0.5
        5 98.6 0.13
        6 99.0 0.07
    };
    
    \addplot[
        fill=violet!50,
        draw=black,
        point meta=y,
        every node near coord/.style={inner ysep=5pt},
        error bars/.cd,
            y dir=both,
            y explicit
    ] 
    table [y error=error] {
        x y error
        1 95.7 0.48
        2 98.0 0.17
        3 98.9 0.21
        4 98.8 0.15
        5 98.9 0.11
        6 99.0 0.1
    };
    
    \draw ({rel axis cs:0,0}|-{axis cs:0,0}) -- ({rel axis cs:1,0}|-{axis cs:0,0});
    \legend{MAC, DAFT MAC}
    \end{axis}
    \end{tikzpicture}
  \vspace{-0.5cm}
  \caption{
    Comparison of CLEVR mean accuracy and 95\% confidence interval ($N=5$) between MAC and DAFT MAC with varying reasoning steps.
  }
  \vspace{-0.5cm}
  \label{fig:step_performance}
\end{wrapfigure}
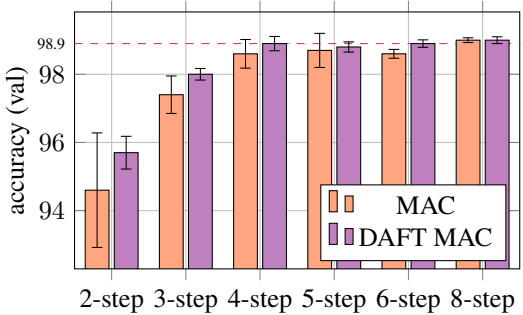

We re-implemented MAC along with DAFT MAC.
We consider a wide range of numbers of steps between $2$ and $30$,
and trained each pair of (method, step number) five times using different random seeds for thorough verification.
As shown in \cref{fig:step_performance}, the accuracy of DAFT MAC outperforms that of the original MAC for fewer reasoning steps (2 $\sim$ 6), and the two methods are roughly tied for larger reasoning steps.
\cite{hudson2018compositional} reported that MAC achieves its best accuracy (98.9\%) at step size $12$;
DAFT MAC reaches equal performance with step size $4$.
In our experiments, MAC and DAFT MAC both reach 99.0\% accuracy at step size $8$.
Increasing step size beyond $8$ results in practically the same performance while requiring more computation; 
in our experiments, 12-step took $\sim$28\% more time compared to 8-step.

The fact that the accuracy of DAFT MAC does not increase when increasing the reasoning step beyond four suggests that four reasoning steps are sufficient for the CLEVR dataset.
We provide more justification for this claim in \cref{subsec:measure} by quantifying the effective number of reasoning steps in each model.

\begin{table}[ht]\centering
  \fontsize{9pt}{9pt}\selectfont
  \caption{Run-time analysis of MAC and DAFT MAC with various ODE solvers.}
  \begin{tabular}{lc|ccc}
  \toprule
  Model & MAC & DAFT MAC & DAFT MAC & DAFT MAC \\
  Solver & - & Euler & Runge-Kutta 4th order & Dormand-Prince \\
  \midrule
  Accuracy & $98.6 \pm 0.2$ & $98.7 \pm 0.2$ & $98.9 \pm 0.2$ & $98.9 \pm 0.2$ \\
  TLT & $2.06 \pm 0.15$ & $1.76 \pm 0.07$ & $1.62 \pm 0.06$ & $1.62 \pm 0.06$ \\
  Time (ms) & $153.7 \pm 3.8$ (1x) & $167.9 \pm 1.7$ (1.09x) & $189.7 \pm 1.9$ (1.23x) & $365.5 \pm 12.5$ (2.37x) \\
\bottomrule
\end{tabular}
\end{table}

We additionally ran a more detailed run-time analysis. 
We measured the accuracy, TLT, and time for inferring a batch of $64$ question-image pairs, using various ODE solvers \textit{during evaluation} of five different 4-step DAFT MAC.
We used two fixed-step solvers (Euler method and Runge-Kutta 4th order method with 3/8 rule) and one adaptive-step solver (Dormand-Prince method) that we used during training.
We found that during evaluation, Runge-Kutta solves all the dynamics generated from CLEVR dataset.
Note that even the simplest Euler method results in higher accuracy and lower TLT compared to vanila MAC.

\subsection{Interpretability} \label{subsec:interpretability} \begin{figure}[ht]
  \centering
  \input{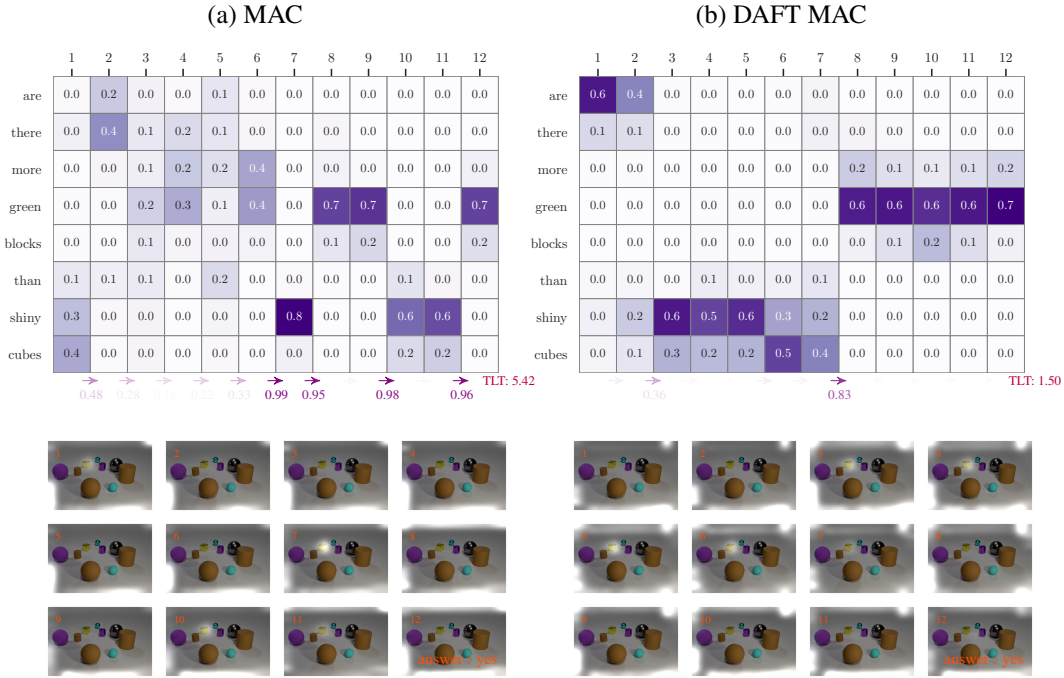}
  \caption{
    Attention maps for the question \textit{"Are there more green blocks than shiny cubes?"} and its accompanying image, the same data used to show attention logit map in \cref{fig:daft}.
    (a) and (b) shows the actual softmax-ed textual and visual attention map which used to acquire the control vector and the information vector in MAC and DAFT MAC, respectively.
  }
  \label{fig:interpretation}
\end{figure}

Many attention-based machine reasoning models put emphasis on the interpretability of the attention map \citep{lu2016hierarchical, kim2018bilinear, hudson2018compositional}.
Indeed, the attention map is a great source of interpretation since it points to specific temporal and spatial points helping our mind to interpret the observation.
In \cref{fig:interpretation}, we compared the qualitative visualization of attention maps for MAC and DAFT MAC.
One can see that DAFT's human prior is beneficial for interpretation in several ways: 

\paragraph{Chunking}
Compared to MAC, DAFT MAC produces more clustered and chunky attention maps.
The question \textit{"Are there more green blocks than shiny cubes?"} contains two noun phrases (NP), \textit{more green blocks} and \textit{shiny cubes}, when parsed to \texttt{(S Are there (NP (ADJP (ADVP more) green) blocks) (PP than (NP shiny cubes)))}.
In this simple case, an ideal solver would only see each NP once to solve the problem.
In \cref{fig:interpretation}, MAC distributes its attention to multiple temporally distant position to retrieve information while DAFT MAC distributes its attention to the chunks which are the same number as the question's NPs.

\paragraph{Consistency}
The attention maps produced by DAFT MAC presents a consistent progression of focus.
We observed that DAFT MACs initialized with different seeds shares the order of transition.
While the learned attention map of MAC varies greatly across different initializations, DAFT MAC consistently attends to \textit{shiny cube} first and then afterwards to \textit{more green blocks} (see \cref{fig:stability_mac} and \cref{fig:stability_daftmac} in the appendix for the clear distinction).

\paragraph{Interpolation}
Since the solution of IVP can yield an attention map for any given point in time, we can easily interpolate the attention maps in-between two adjacent steps.
See \cref{fig:interpolation} in the appendix for a visualization of these interpolated maps.
Note that although we visualized the interpolation with the sampling rate of 20 due to limited space, this rate can go infinitely high since DAFT is continuous in time.
This interpolation differs from simple linear interpolation since DAFT has non-linear dynamics.
\subsection{Total Length of Transition} \label{subsec:measure} To mesure the description length of a given attention map, we first define the \textit{length} of the map.
Recall that the attention map is a categorical distribution over input tokens.
A simple example of quantifying the distance of such a map is to choose the word to which the model focused on most at each time step, and measure the number of times this shifted.
For example, the attention map of DAFT MAC in \cref{fig:interpretation} can be simplified as \texttt{["are", "shiny", "cubes", "green"]} and the map of MAC as \texttt{["cubes", "there", "green, "than", "green", "shiny", "green", "shiny", "green"]}.
If we measure the length in this way, the lengths become $4$ and $9$, respectively.

However, since we have more finer information than just gathering tokens with maximum values, we can employ probabilistic measures of distances.
The distances will generally follow the simple discrete measurement and can measure more precise length of given attention map.
Thus we use the Jensen-Shannon divergence \citep{lin1991divergence} 
to measure the amount of shift between attention maps throughout reasoning.
We chose the Jensen-Shannon divergence because it is bounded ($\textrm{JSD}(P || Q) \in [0, 1]$).

\begin{definition}{Length of Transition (LT)}\\
Let $\mathbf{p}_t \in \mathbb{R}^S$ be the attention probability for time $t = 1, \ldots, T$.
The Length of Transition (LT) at time $t$ is defined as:%
\be
\text{LT}(t) 
= \textrm{JSD}(\mathbf{p}_t || \mathbf{p}_{t+1})
= \frac{1}{2} \sum_{s=1}^{S} p_{t}^s \cdot \log_2 \frac{2 \cdot p_{t}^s}{p_{t}^s + p_{t+1}^s} + p_{t+1}^s \cdot \log_2 \frac{2 \cdot p_{t+1}^s}{p_{t}^s + p_{t+1}^s}
\ee
where $p_t^s$ is the $s$-th element of $\mathbf{p}_t$.
\label{def:lt}
\end{definition}

We further define total length of transition (TLT) as $\text{TLT} = \sum_{i=1}^{T} LT(i)$
\footnote{This is quite similar to the length of the prequential (online) code of \cite{blier2018description}, with the difference that theirs is a sum of negative log probabilities instead of a JS divegence}.
In default, TLT is bounded by $T-1$, and if TLT considers $LT(0)$, it is bounded by $T$.
One can concatenate uniformly distributed attention to $\mathbf{a}$ as a starting attention $\mathbf{a}_0$ to get $LT(0)$.
We do not use $LT(0)$ when calculating TLT throughout this paper, making it bounded by $T-1$.
Furthermore, we argue that a model with low TLT is more likely to produce consistent attention maps across different initializations since TLT imposes an upper bound on the amount the model's attention can change.
We denoted LTs and TLT for MAC and DAFT MAC at the below of attention maps in \cref{fig:interpretation}.

\begin{figure}[ht]
    \centering
    \begin{tikzpicture}
    \begin{axis}[
        height=5cm,
        width=14cm,
        ymin=0,
        /pgf/bar width=9pt,%
        xtick={1,...,10},
        xticklabels={%
            2-step,
            3-step,
            4-step,
            5-step,
            6-step,
            8-step,
            12-step,
            16-step,
            20-step,
            30-step},
        ylabel={\small Total Length of Transition},
        y label style={at={(0.03,0.5)}},
        grid=major,
        ybar,
        legend pos=north west,
        legend style={nodes={scale=0.8, transform shape}},
        legend image post style={scale=0.8},
        ]
    
    \addplot[
        fill=orange!50,
        draw=black,
        point meta=y,
        every node near coord/.style={inner ysep=5pt},
        error bars/.cd,
            y dir=both,
            y explicit
    ] 
    table [y error=error] {
        x y error
        1 0.83 0.08
        2 1.48 0.22
        3 2.06 0.15
        4 2.77 0.14
        5 3.00 0.22
        6 4.32 0.39
        7 6.69 0.21
        8 7.6 0.41
        9 9.48 1.01
        10 14.28 0.87
    };

    \addplot[
        fill=violet!50,
        draw=black,
        point meta=y,
        every node near coord/.style={inner ysep=5pt},
        error bars/.cd,
            y dir=both,
            y explicit
    ] 
    table [y error=error] {
        x y error
        1 0.83 0.06
        2 1.13 0.04
        3 1.62 0.06
        4 1.72 0.12
        5 1.91 0.09
        6 2.25 0.2
        7 2.82 0.54
        8 3.86 0.29
        9 3.84 0.4
        10 4.54 0.62
    };

    \draw ({rel axis cs:0,0}|-{axis cs:0,0}) -- ({rel axis cs:1,0}|-{axis cs:0,0});
    \legend{MAC, DAFT MAC}
    \end{axis}
    \end{tikzpicture}
    \caption{
        Comparison of CLEVR mean TLT and its 95\% confidence interval ($N=5$) between MAC and DAFT MAC with varying reasoning steps.
    }
    \label{fig:step_shifts}
\end{figure}
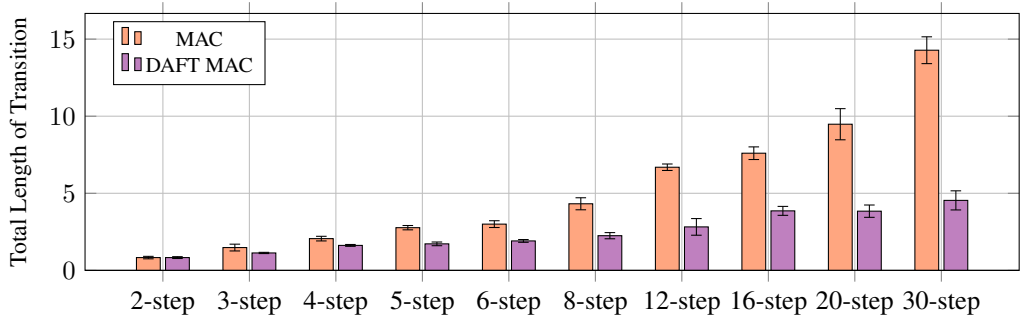

\cref{fig:step_shifts} shows the TLT values of MAC and DAFT MAC.
When the number of reasoning steps increases, the TLT of DAFT MAC is relatively unchanged while that of MAC increases with step number.
This result supports the qualitative result shown before and demonstrates that DAFT MAC consistently results in simplified reasoning paths across the whole dataset, rather than only in a few cherry-picked examples.
In \cref{subsec:performance}, we have argued that the 4-step is enough for solving CLEVR.
In \cref{fig:step_shifts}, one can see that step-wise growth reaches its maximum in 4-step (for clear view, see \cref{fig:growth_tlt} in the appendix),
implying that the model requires more space to navigate its focus when the step size is smaller than four.

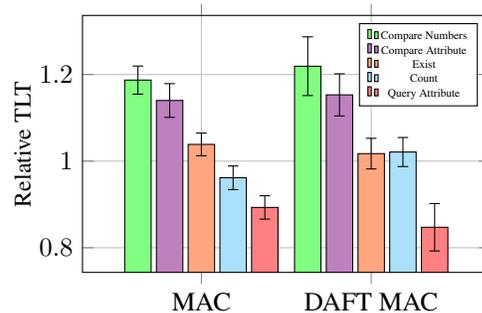
\begin{wrapfigure}{R}{0.5\textwidth}
    \centering
    \vspace{-0.2cm}
    \begin{tikzpicture}
    \begin{axis}[
        height=5cm,
        width=7cm,
        /pgf/bar width=10pt,%
        xtick={1,2},
        xticklabels={
            \shortstack{MAC},
            \shortstack{DAFT MAC},
        },
        enlarge x limits=0.7,
        ylabel={\small Relative TLT},
        y label style={at={(0.08,0.5)}},
        grid=major,
        ybar,
        legend pos=north east,
        legend style={nodes={scale=0.43, transform shape}},
        legend image post style={scale=0.43},
        ]

    \addplot[
        fill=green!50,
        draw=black,
        point meta=y,
        every node near coord/.style={inner ysep=5pt},
        error bars/.cd,
            y dir=both,
            y explicit
    ] 
    table [y error=error] {
        x y error
        1 1.1871 0.0324
        2 1.2191 0.0678
    };

    \addplot[
        fill=violet!50,
        draw=black,
        point meta=y,
        every node near coord/.style={inner ysep=5pt},
        error bars/.cd,
            y dir=both,
            y explicit
    ] 
    table [y error=error] {
        x y error
        1 1.1400 0.0389
        2 1.1530 0.0486
    };

    \addplot[
        fill=orange!50,
        draw=black,
        point meta=y,
        every node near coord/.style={inner ysep=5pt},
        error bars/.cd,
            y dir=both,
            y explicit
    ] 
    table [y error=error] {
        x y error
        1 1.0386 0.0262
        2 1.0173 0.0354
    };

    \addplot[
        fill=blue!50,
        draw=black,
        point meta=y,
        every node near coord/.style={inner ysep=5pt},
        error bars/.cd,
            y dir=both,
            y explicit
    ] 
    table [y error=error] {
        x y error
        1 0.9616 0.0273
        2 1.0210 0.0335
    };

    \addplot[
        fill=red!50,
        draw=black,
        point meta=y,
        every node near coord/.style={inner ysep=5pt},
        error bars/.cd,
            y dir=both,
            y explicit
    ] 
    table [y error=error] {
        x y error
        1 0.8931 0.0270
        2 0.8473 0.0547
    };
    
    \draw ({rel axis cs:0,0}|-{axis cs:0,0}) -- ({rel axis cs:1,0}|-{axis cs:0,0});
    \legend{Compare Numbers, Compare Attribute, Exist, Count, Query Attribute}
    \end{axis}
    \end{tikzpicture}
    \vspace{-0.2cm}
    \caption{
        Comparison of relative TLT mean accuracy and its 95\% confidence interval ($N=50$) with varying question type.
    }
    \vspace{-0.5cm}
    \label{fig:step_shifts_qtype}
\end{wrapfigure}

\cref{fig:step_shifts_qtype} shows how much TLT each question type yields.
Since TLT grows with the size of the reasoning step, we employed a relative value of TLT to normalize this value across different numbers of training steps.
Relative TLT is defined as $\nicefrac{TLT_t(question\_type)}{TLT_t}$, where $t$ ranges over steps in \cref{fig:step_shifts}.
The fact that each question type's relative TLT has the same order within both MAC and DAFT MAC substantiates TLT's ability to measure reasoning complexity regardless of the specific architecture.

Question types \textit{Compare Numbers} and \textit{Compare Attribute} had higher TLT than other question types.
This is expected since such comparative questions involve more NP chunks than other question types.
When we shrank the step size from four to two, the accuracy of \textit{Query Attribute} question type was pretty much unharmed (99.6 $\rightarrow$ 99.3 in DAFT MAC and 99.6 $\rightarrow$ 97.5 in MAC) while that of other question types significantly dropped.
This is supported by the fact that \textit{Query Attribute} question type had lowest TLT, meaning the question type is solvable using a small number of steps.

\section{Conclusion} \label{sec:conclusion}
We have proposed Dynamics of Attention for Focus Transition (DAFT), which embeds the human prior of continuous focus transition.
In contrast to previous approaches, DAFT learns the dynamics in-between reasoning steps, yielding more interpretable attention maps.
When applied to MAC, the state-of-the-art among models that only use natural supervision, DAFT achieves the same performance while using $\nicefrac{1}{3}$ the number of reasoning steps.
In addition, we proposed a novel metric called Total Length Transition (TLT).
Following the minimum description length principle, TLT measures how good the model is on planning effective, short reasoning path (latent program), which is directly related to the interpretability of the model.

Next on our agenda includes
(1) extending DAFT to other tasks where performance and interpretability are both important to develop a method to balance between the two criteria, and 
(2) investigating what other values TLT can serve as a proxy for.

\clearpage
\bibliographystyle{plainnat}
\bibliography{main.bib}

\begin{thebibliography}{42}
\providecommand{\natexlab}[1]{#1}
\providecommand{\url}[1]{\texttt{#1}}
\expandafter\ifx\csname urlstyle\endcsname\relax
  \providecommand{\doi}[1]{doi: #1}\else
  \providecommand{\doi}{doi: \begingroup \urlstyle{rm}\Url}\fi

\bibitem[Agrawal et~al.(2015)Agrawal, Lu, Antol, Mitchell, Zitnick, Batra, and
  Parikh]{agrawal2015vqa}
Aishwarya Agrawal, Jiasen Lu, Stanislaw Antol, Margaret Mitchell, C~Lawrence
  Zitnick, Dhruv Batra, and Devi Parikh.
\newblock Vqa: Visual question answering.
\newblock \emph{arXiv preprint arXiv:1505.00468}, 2015.

\bibitem[Andreas et~al.(2016)Andreas, Rohrbach, Darrell, and
  Klein]{andreas2016neural}
Jacob Andreas, Marcus Rohrbach, Trevor Darrell, and Dan Klein.
\newblock Neural module networks.
\newblock In \emph{Proceedings of the IEEE Conference on Computer Vision and
  Pattern Recognition}, pages 39--48, 2016.

\bibitem[Blier and Ollivier(2018)]{blier2018description}
L{\'e}onard Blier and Yann Ollivier.
\newblock The description length of deep learning models.
\newblock In \emph{Advances in Neural Information Processing Systems}, pages
  2216--2226, 2018.

\bibitem[Bottou(2014)]{bottou2014machine}
L{\'e}on Bottou.
\newblock From machine learning to machine reasoning.
\newblock \emph{Machine learning}, 94\penalty0 (2):\penalty0 133--149, 2014.

\bibitem[Chen et~al.(2018)Chen, Rubanova, Bettencourt, and
  Duvenaud]{chen2018neural}
Tian~Qi Chen, Yulia Rubanova, Jesse Bettencourt, and David~K Duvenaud.
\newblock Neural ordinary differential equations.
\newblock In \emph{Advances in Neural Information Processing Systems}, pages
  6571--6583, 2018.

\bibitem[Dehghani et~al.(2018)Dehghani, Gouws, Vinyals, Uszkoreit, and
  Kaiser]{dehghani2018universal}
Mostafa Dehghani, Stephan Gouws, Oriol Vinyals, Jakob Uszkoreit, and {\L}ukasz
  Kaiser.
\newblock Universal transformers.
\newblock \emph{arXiv preprint arXiv:1807.03819}, 2018.

\bibitem[Dormand and Prince(1980)]{dormand1980family}
John~R Dormand and Peter~J Prince.
\newblock A family of embedded runge-kutta formulae.
\newblock \emph{Journal of computational and applied mathematics}, 6\penalty0
  (1):\penalty0 19--26, 1980.

\bibitem[Dupont et~al.(2019)Dupont, Doucet, and Teh]{dupont2019augmented}
Emilien Dupont, Arnaud Doucet, and Yee~Whye Teh.
\newblock Augmented neural odes.
\newblock \emph{arXiv preprint arXiv:1904.01681}, 2019.

\bibitem[Feldman(2009)]{feldman2009bayes}
Jacob Feldman.
\newblock Bayes and the simplicity principle in perception.
\newblock \emph{Psychological review}, 116\penalty0 (4):\penalty0 875, 2009.

\bibitem[Feldman(2016)]{feldman2016simplicity}
Jacob Feldman.
\newblock The simplicity principle in perception and cognition.
\newblock \emph{Wiley Interdisciplinary Reviews: Cognitive Science}, 7\penalty0
  (5):\penalty0 330--340, 2016.

\bibitem[Glorot and Bengio(2010)]{glorot2010understanding}
Xavier Glorot and Yoshua Bengio.
\newblock Understanding the difficulty of training deep feedforward neural
  networks.
\newblock In \emph{Proceedings of the thirteenth international conference on
  artificial intelligence and statistics}, pages 249--256, 2010.

\bibitem[Goyal et~al.(2017)Goyal, Khot, Summers-Stay, Batra, and
  Parikh]{goyal2017making}
Yash Goyal, Tejas Khot, Douglas Summers-Stay, Dhruv Batra, and Devi Parikh.
\newblock Making the v in vqa matter: Elevating the role of image understanding
  in visual question answering.
\newblock In \emph{Proceedings of the IEEE Conference on Computer Vision and
  Pattern Recognition}, pages 6904--6913, 2017.

\bibitem[Graves(2016)]{graves2016adaptive}
Alex Graves.
\newblock Adaptive computation time for recurrent neural networks.
\newblock \emph{arXiv preprint arXiv:1603.08983}, 2016.

\bibitem[Haber and Ruthotto(2017)]{haber2017stable}
Eldad Haber and Lars Ruthotto.
\newblock Stable architectures for deep neural networks.
\newblock \emph{Inverse Problems}, 34\penalty0 (1):\penalty0 014004, 2017.

\bibitem[He et~al.(2017)He, Gkioxari, Doll{\'a}r, and Girshick]{he2017mask}
Kaiming He, Georgia Gkioxari, Piotr Doll{\'a}r, and Ross Girshick.
\newblock Mask r-cnn.
\newblock In \emph{Proceedings of the IEEE international conference on computer
  vision}, pages 2961--2969, 2017.

\bibitem[Hochberg and McAlister(1953)]{hochberg1953quantitative}
Julian Hochberg and Edward McAlister.
\newblock A quantitative approach, to figural" goodness".
\newblock \emph{Journal of Experimental Psychology}, 46\penalty0 (5):\penalty0
  361, 1953.

\bibitem[Hu et~al.(2017)Hu, Andreas, Rohrbach, Darrell, and
  Saenko]{hu2017learning}
Ronghang Hu, Jacob Andreas, Marcus Rohrbach, Trevor Darrell, and Kate Saenko.
\newblock Learning to reason: End-to-end module networks for visual question
  answering.
\newblock In \emph{Proceedings of the IEEE International Conference on Computer
  Vision}, pages 804--813, 2017.

\bibitem[Hudson and Manning(2018)]{hudson2018compositional}
Drew~A Hudson and Christopher~D Manning.
\newblock Compositional attention networks for machine reasoning.
\newblock \emph{arXiv preprint arXiv:1803.03067}, 2018.

\bibitem[Hudson and Manning(2019)]{hudson2019gqa}
Drew~A Hudson and Christopher~D Manning.
\newblock Gqa: a new dataset for compositional question answering over
  real-world images.
\newblock \emph{arXiv preprint arXiv:1902.09506}, 2019.

\bibitem[Ilyas et~al.(2019)Ilyas, Santurkar, Tsipras, Engstrom, Tran, and
  Madry]{ilyas2019adversarial}
Andrew Ilyas, Shibani Santurkar, Dimitris Tsipras, Logan Engstrom, Brandon
  Tran, and Aleksander Madry.
\newblock Adversarial examples are not bugs, they are features.
\newblock \emph{arXiv preprint arXiv:1905.02175}, 2019.

\bibitem[Johnson et~al.(2017{\natexlab{a}})Johnson, Hariharan, van~der Maaten,
  Fei-Fei, Lawrence~Zitnick, and Girshick]{johnson2017clevr}
Justin Johnson, Bharath Hariharan, Laurens van~der Maaten, Li~Fei-Fei,
  C~Lawrence~Zitnick, and Ross Girshick.
\newblock Clevr: A diagnostic dataset for compositional language and elementary
  visual reasoning.
\newblock In \emph{Proceedings of the IEEE Conference on Computer Vision and
  Pattern Recognition}, pages 2901--2910, 2017{\natexlab{a}}.

\bibitem[Johnson et~al.(2017{\natexlab{b}})Johnson, Hariharan, van~der Maaten,
  Hoffman, Fei-Fei, Lawrence~Zitnick, and Girshick]{johnson2017inferring}
Justin Johnson, Bharath Hariharan, Laurens van~der Maaten, Judy Hoffman,
  Li~Fei-Fei, C~Lawrence~Zitnick, and Ross Girshick.
\newblock Inferring and executing programs for visual reasoning.
\newblock In \emph{Proceedings of the IEEE International Conference on Computer
  Vision}, pages 2989--2998, 2017{\natexlab{b}}.

\bibitem[Kim et~al.(2018)Kim, Jun, and Zhang]{kim2018bilinear}
Jin-Hwa Kim, Jaehyun Jun, and Byoung-Tak Zhang.
\newblock Bilinear attention networks.
\newblock In \emph{Advances in Neural Information Processing Systems}, pages
  1564--1574, 2018.

\bibitem[Kingma and Ba(2014)]{kingma2014adam}
Diederik~P Kingma and Jimmy Ba.
\newblock Adam: A method for stochastic optimization.
\newblock \emph{arXiv preprint arXiv:1412.6980}, 2014.

\bibitem[Kolmogorov(1963)]{kolmogorov1963tables}
Andrei~N Kolmogorov.
\newblock On tables of random numbers.
\newblock \emph{Sankhy{\=a}: The Indian Journal of Statistics, Series A}, pages
  369--376, 1963.

\bibitem[Krishna et~al.(2017)Krishna, Zhu, Groth, Johnson, Hata, Kravitz, Chen,
  Kalantidis, Li, Shamma, et~al.]{krishna2017visual}
Ranjay Krishna, Yuke Zhu, Oliver Groth, Justin Johnson, Kenji Hata, Joshua
  Kravitz, Stephanie Chen, Yannis Kalantidis, Li-Jia Li, David~A Shamma, et~al.
\newblock Visual genome: Connecting language and vision using crowdsourced
  dense image annotations.
\newblock \emph{International Journal of Computer Vision}, 123\penalty0
  (1):\penalty0 32--73, 2017.

\bibitem[Lage et~al.(2018)Lage, Ross, Gershman, Kim, and
  Doshi-Velez]{lage2018human}
Isaac Lage, Andrew Ross, Samuel~J Gershman, Been Kim, and Finale Doshi-Velez.
\newblock Human-in-the-loop interpretability prior.
\newblock In \emph{Advances in Neural Information Processing Systems}, pages
  10159--10168, 2018.

\bibitem[Lin(1991)]{lin1991divergence}
Jianhua Lin.
\newblock Divergence measures based on the shannon entropy.
\newblock \emph{IEEE Transactions on Information theory}, 37\penalty0
  (1):\penalty0 145--151, 1991.

\bibitem[Lu et~al.(2016)Lu, Yang, Batra, and Parikh]{lu2016hierarchical}
Jiasen Lu, Jianwei Yang, Dhruv Batra, and Devi Parikh.
\newblock Hierarchical question-image co-attention for visual question
  answering.
\newblock In \emph{Advances In Neural Information Processing Systems}, pages
  289--297, 2016.

\bibitem[Lu et~al.(2017)Lu, Zhong, Li, and Dong]{lu2017beyond}
Yiping Lu, Aoxiao Zhong, Quanzheng Li, and Bin Dong.
\newblock Beyond finite layer neural networks: Bridging deep architectures and
  numerical differential equations.
\newblock \emph{arXiv preprint arXiv:1710.10121}, 2017.

\bibitem[Mao et~al.(2018)Mao, Gan, Kohli, Tenenbaum, and Wu]{mao2018neuro}
Jiayuan Mao, Chuang Gan, Pushmeet Kohli, Joshua~B Tenenbaum, and Jiajun Wu.
\newblock The neuro-symbolic concept learner: Interpreting scenes, words, and
  sentences from natural supervision.
\newblock 2018.

\bibitem[Mascharka et~al.(2018)Mascharka, Tran, Soklaski, and
  Majumdar]{mascharka2018transparency}
David Mascharka, Philip Tran, Ryan Soklaski, and Arjun Majumdar.
\newblock Transparency by design: Closing the gap between performance and
  interpretability in visual reasoning.
\newblock In \emph{Proceedings of the IEEE conference on computer vision and
  pattern recognition}, pages 4942--4950, 2018.

\bibitem[Paszke et~al.(2017)Paszke, Gross, Chintala, Chanan, Yang, DeVito, Lin,
  Desmaison, Antiga, and Lerer]{paszke2017automatic}
Adam Paszke, Sam Gross, Soumith Chintala, Gregory Chanan, Edward Yang, Zachary
  DeVito, Zeming Lin, Alban Desmaison, Luca Antiga, and Adam Lerer.
\newblock Automatic differentiation in pytorch.
\newblock In \emph{NIPS-W}, 2017.

\bibitem[Perez et~al.(2018)Perez, Strub, De~Vries, Dumoulin, and
  Courville]{perez2018film}
Ethan Perez, Florian Strub, Harm De~Vries, Vincent Dumoulin, and Aaron
  Courville.
\newblock Film: Visual reasoning with a general conditioning layer.
\newblock In \emph{Thirty-Second AAAI Conference on Artificial Intelligence},
  2018.

\bibitem[Poursabzi-Sangdeh et~al.(2018)Poursabzi-Sangdeh, Goldstein, Hofman,
  Vaughan, and Wallach]{poursabzi2018manipulating}
Forough Poursabzi-Sangdeh, Daniel~G Goldstein, Jake~M Hofman, Jennifer~Wortman
  Vaughan, and Hanna Wallach.
\newblock Manipulating and measuring model interpretability.
\newblock \emph{arXiv preprint arXiv:1802.07810}, 2018.

\bibitem[Rissanen(1978)]{rissanen1978modeling}
Jorma Rissanen.
\newblock Modeling by shortest data description.
\newblock \emph{Automatica}, 14\penalty0 (5):\penalty0 465--471, 1978.

\bibitem[Ruthotto and Haber(2018)]{ruthotto2018deep}
Lars Ruthotto and Eldad Haber.
\newblock Deep neural networks motivated by partial differential equations.
\newblock \emph{arXiv preprint arXiv:1804.04272}, 2018.

\bibitem[Santoro et~al.(2017)Santoro, Raposo, Barrett, Malinowski, Pascanu,
  Battaglia, and Lillicrap]{santoro2017simple}
Adam Santoro, David Raposo, David~G Barrett, Mateusz Malinowski, Razvan
  Pascanu, Peter Battaglia, and Timothy Lillicrap.
\newblock A simple neural network module for relational reasoning.
\newblock In \emph{Advances in neural information processing systems}, pages
  4967--4976, 2017.

\bibitem[Suarez et~al.(2018)Suarez, Johnson, and Li]{suarez2018ddrprog}
Joseph Suarez, Justin Johnson, and Fei-Fei Li.
\newblock Ddrprog: A clevr differentiable dynamic reasoning programmer.
\newblock \emph{arXiv preprint arXiv:1803.11361}, 2018.

\bibitem[Vendetti and Bunge(2014)]{vendetti2014evolutionary}
Michael~S Vendetti and Silvia~A Bunge.
\newblock Evolutionary and developmental changes in the lateral frontoparietal
  network: a little goes a long way for higher-level cognition.
\newblock \emph{Neuron}, 84\penalty0 (5):\penalty0 906--917, 2014.

\bibitem[Xiong et~al.(2016)Xiong, Merity, and Socher]{xiong2016dynamic}
Caiming Xiong, Stephen Merity, and Richard Socher.
\newblock Dynamic memory networks for visual and textual question answering.
\newblock In \emph{International conference on machine learning}, pages
  2397--2406, 2016.

\bibitem[Yi et~al.(2018)Yi, Wu, Gan, Torralba, Kohli, and
  Tenenbaum]{yi2018neural}
Kexin Yi, Jiajun Wu, Chuang Gan, Antonio Torralba, Pushmeet Kohli, and Josh
  Tenenbaum.
\newblock Neural-symbolic vqa: Disentangling reasoning from vision and language
  understanding.
\newblock In \emph{Advances in Neural Information Processing Systems}, pages
  1031--1042, 2018.

\end{thebibliography}
\clearpage
\appendix \section{Reasoning with Additional Supervision} \label{app:related}
In this section, we detail the additional supervisions and taxonomize previous studies on visual reasoning according to each additional supervision signal used.

\paragraph{Program}
A (functional) program is a set of logical functions that can be executed on an image's scene graph.
Programs are a valuable supervision for VQA models since it enables the model to convert a natural language question into excutable functions \citep{andreas2016neural, hu2017learning, mascharka2018transparency, yi2018neural}.
Such composition of functions provides far better interpretability than that of models which interact with raw sensory data (i.e., natural supervisions).

However, since programs are generally disentangled from scene graphs, their supervision can keep the model from learning to skip unnecessary steps as we discussed in the introduction.
This problem can be handled by only giving the model \textit{optimal} programs.
Synthetically generated dataset such as CLEVR \citep{johnson2017clevr} pruned out suboptimal programs like shown in \cref{fig:clevr_example} by inspecting their related scene graph.
However, even such pruning requires a perfect scene graph, which is not available in real-world datasets.
Therefore the programs provided in real-world VQA datasets such as GQA \citep{hudson2019gqa} are inherently suboptimal since they are generated from approximate scene graphs.

\paragraph{Object Mask}
The other thing that makes solving VQA hard is that each node in the scene graph corresponds to a different group of pixels.
Object masks help the model locate nodes (i.e., objects) of the scene graph before the reasoning steps (i.e., program execution).
Well annotated masks relieve the model from finding objects and allow them to solely concentrate on reasoning.

\cite{yi2018neural} generated object masks for the CLEVR dataset using its scene graph, and used them as supervisory signals for inferring the scene graph.
In the specific setup of CLEVR, they achieved near perfect performance (99.8\%).
Such result shows that object masks and their corresponding inference module such as Mask R-CNN \citep{he2017mask} are enough to build an exact scene graph in synthetically generated images.
However, annotating object masks and their scene graphs for real-world images is still an open problem.

\section{Implementation Details} \label{app:impl} To solve the initial value problem of the ODE, we used \texttt{torchdiffeq} \citep{chen2018neural}.
For designing and accelerating the computation graph of the model, we used \texttt{pytorch} 1.0.1 \citep{paszke2017automatic} with CUDA 9.2 on an Nvidia V100 GPU.
Every experiment was performed with five different initial seeds by fixing the inital seed with \texttt{manual\_seed()} for \texttt{python}, \texttt{pytorch}, and \texttt{numpy}.

We used the \texttt{Adam} optimizer \citep{kingma2014adam} with learning rate 1e-4 for all experiments, and halved the learning rate whenever the validation accuracy stopped improving for more than one epoch.
We trained using batches of 64 training data and terminated training when the learning rate went under 1e-7.
The size of all hidden dimensions was fixed to 512 except for the word embedding layer, which was 300.
All weights for the affine transformation were initialized with \texttt{xavier} initialization \citep{glorot2010understanding}, and word embeddings were initialized to random vectors using a uniform distribution following the settings of MAC.

\section{Experiments on GQA} \label{app:gqa} \begin{figure}[ht]
  \centering
  \input{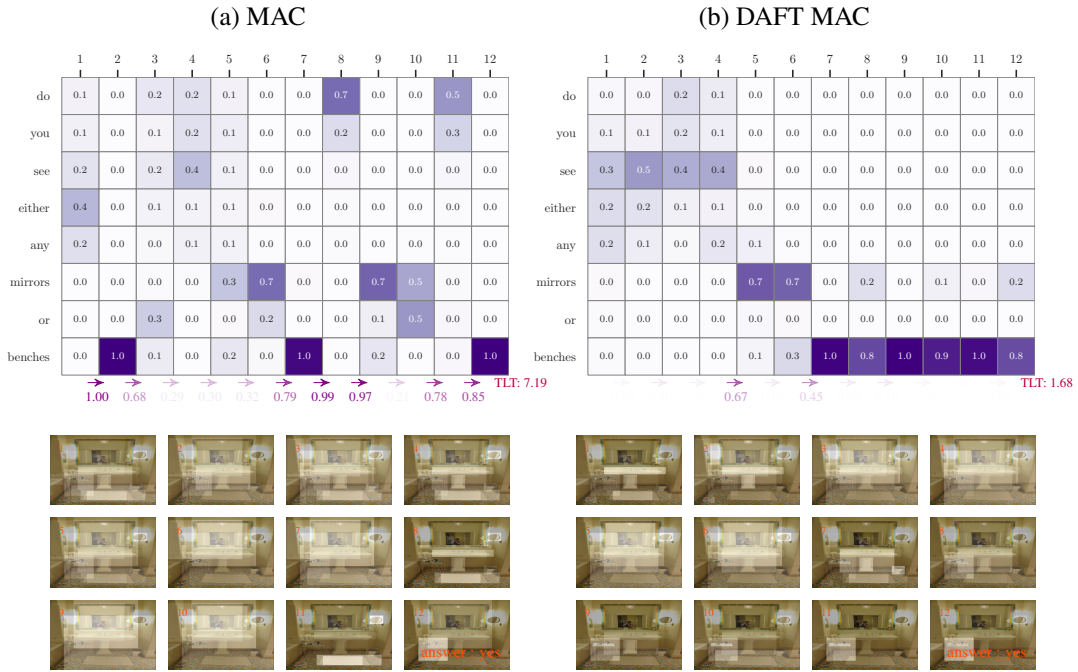}
  \caption{
    A graphical description of how attention maps change in MAC and DAFT MAC for a GQA example.
    The given question is \textit{"do you see either any mirrors or benches?"}.
    Attention maps of 12-step (a) MAC and (b) DAFT MAC are shown.
    In both textual and visual, DAFT MAC's attention changes from mirrors to benches smoothly.
  }
  \label{fig:daft_gqa}
  \end{figure}

\begin{figure}[ht]
  \centering
  \begin{tikzpicture}
    \begin{axis}[
        height=5cm,
        width=10cm,
        ymin=0,
        /pgf/bar width=9pt,%
        xtick={1,...,7},
        xticklabels={%
            2-step,
            3-step,
            4-step,
            5-step,
            6-step,
            8-step,
            12-step
            },
        ylabel={\small Total Length of Transition},
        y label style={at={(0.03,0.5)}},
        grid=major,
        ybar,
        legend pos=north west,
        legend style={nodes={scale=0.8, transform shape}},
        legend image post style={scale=0.8},
        ]
    
    \addplot[
        fill=orange!50,
        draw=black,
        point meta=y,
        every node near coord/.style={inner ysep=5pt},
        error bars/.cd,
            y dir=both,
            y explicit
    ] 
    table [y error=error] {
        x y error
        1 0.35 0.06
        2 0.65 0.11
        3 1.16 0.22
        4 1.62 0.20
        5 1.94 0.23
        6 3.07 0.29
        7 5.09 0.19
    };

    \addplot[
        fill=violet!50,
        draw=black,
        point meta=y,
        every node near coord/.style={inner ysep=5pt},
        error bars/.cd,
            y dir=both,
            y explicit
    ] 
    table [y error=error] {
        x y error
        1 0.20 0.04
        2 0.35 0.08
        3 0.52 0.06
        4 0.59 0.13
        5 0.72 0.08
        6 0.92 0.18
        7 1.39 0.35
    };

    \draw ({rel axis cs:0,0}|-{axis cs:0,0}) -- ({rel axis cs:1,0}|-{axis cs:0,0});
    \legend{MAC, DAFT MAC}
    \end{axis}
    \end{tikzpicture}
  \caption{
      Comparison of GQA mean TLT and its 95\% confidence interval ($N=5$) between MAC and DAFT MAC with varying reasoning steps.
  }
  \label{fig:step_shifts_gqa}
\end{figure}
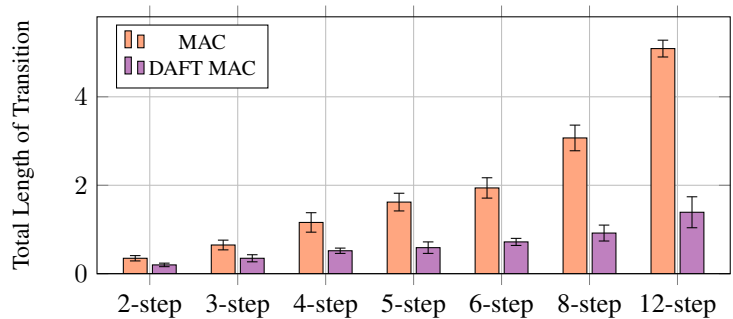

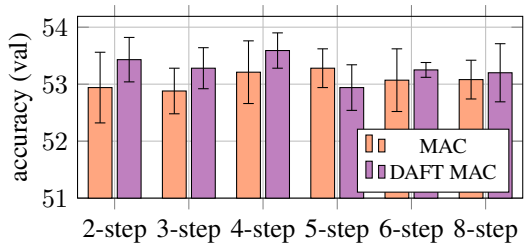
\begin{wrapfigure}{R}{0.5\textwidth}
  \centering
  \begin{tikzpicture}
    \begin{axis}[
        height=4cm,
        width=7.5cm,
        ymin=51,
        /pgf/bar width=9pt,%
        xtick={1,...,10},
        xticklabels={%
            2-step,
            3-step,
            4-step,
            5-step,
            6-step,
            8-step,
            },
        ylabel={accuracy (val)},
        y label style={at={(0.08,0.5)}},
        grid=major,
        ybar,
        legend pos=south east,
        legend style={nodes={scale=0.8, transform shape}},
        legend image post style={scale=0.8},
        ]
    
    \addplot[
        fill=orange!50,
        draw=black,
        point meta=y,
        every node near coord/.style={inner ysep=5pt},
        error bars/.cd,
            y dir=both,
            y explicit
    ] 
    table [y error=error] {
        x y error
        1 52.94 0.62
        2 52.88 0.40
        3 53.21 0.55
        4 53.28 0.34
        5 53.07 0.55
        6 53.08 0.34
    };
    
    \addplot[
        fill=violet!50,
        draw=black,
        point meta=y,
        every node near coord/.style={inner ysep=5pt},
        error bars/.cd,
            y dir=both,
            y explicit
    ] 
    table [y error=error] {
        x y error
        1 53.43 0.39
        2 53.28 0.36
        3 53.59 0.31
        4 52.94 0.40
        5 53.25 0.13
        6 53.20 0.51
    };
    
    \draw ({rel axis cs:0,0}|-{axis cs:0,0}) -- ({rel axis cs:1,0}|-{axis cs:0,0});
    \legend{MAC, DAFT MAC}
    \end{axis}
    \end{tikzpicture}
  \vspace{-0.5cm}
  \caption{
    Comparison of overall GQA mean accuracy and its 95\% confidence interval ($N=5$) between MAC and DAFT MAC with varying reasoning steps.
  }
  \vspace{-0.5cm}
  \label{fig:step_performance_gqa}
\end{wrapfigure}

The GQA \citep{hudson2019gqa} dataset is a real-world VQA dataset where all questions are generated compositionally by injecting scene graph information into question templates.
Although it shares most details with CLEVR dataset, its questions are far less complex than that of CLEVR.
Creating complex questions in GQA is challenging because while each of its images typically contain objects from many different classes, the number of objects for each class is small.

On the other hand, an image in the CLEVR dataset contains few classes of objects, but the number of objects per class is large;
this enables CLEVR to make compositionally complex questions.
In \cref{fig:daft_gqa}, we compare the attention maps of MAC and DAFT MAC just as we did in \cref{fig:interpretation}.

\cref{fig:step_shifts_gqa} shows the TLTs of the two methods on the GQA dataset while varying reasoning steps.
Compared with \cref{fig:step_shifts}, one can see that the TLT of GQA is far less than that of CLEVR.
This confirms that questions of GQA is indeed less complex than that of CLEVR.

\cref{fig:step_performance_gqa} shows accuracies of MAC and DAFT MAC when evaulated on the GQA dataset.
As shown in the figure, there is not much difference in accuracy across step sizes for both MAC and DAFT MAC.
It tells us that 2-step is roughly enough, as 4-step MAC in CLEVR did.
We would like to note that we observed many runs that achieve over 54\% validation accuracy (which matches with the accuracy \cite{hudson2019gqa} reported) for some periods.
However, the accuracies fall shortly after the peak and converged in reported accuracies.
Since we reported all accuracies and TLTs with the model which passed through full training session as mentioned in \cref{app:impl} throughout the paper,
we report the results of GQA in the same manner.

\section{Additional Figures}

\begin{figure}[ht]
    \centering
    \includegraphics[width=1.0\textwidth]{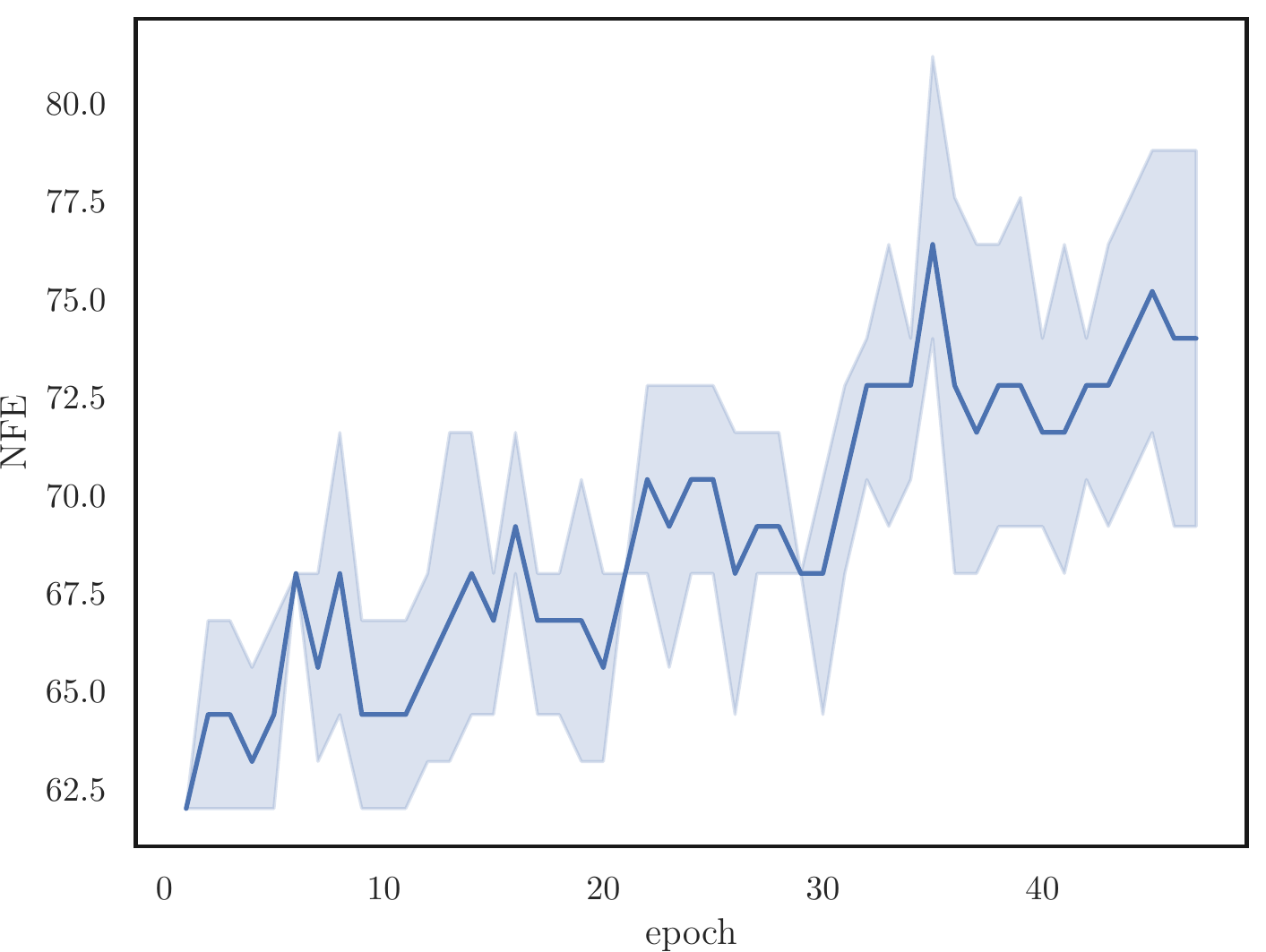}
    \caption{
        Growth of the Number of Function Evaluation (NFE) for 4-step DAFT MAC as training progresses. 
        Mean value and 95\% confidence interval ($N=5$) are denoted as line and gradation.
    }
    \label{fig:nfe}
\end{figure}
    
\begin{figure}[ht]
    \centering
    \begin{tikzpicture}
    \begin{axis}[
        height=5cm,
        width=14cm,
        /pgf/bar width=9pt,%
        xtick={1,...,9},
        xticklabels={%
            3,
            4,
            5,
            6,
            8,
            12,
            16,
            20,
            30,
            },
        ylabel={\small Mean growth of TLT},
        y label style={at={(0.03,0.5)}},
        grid=major,
        ybar,
        legend pos=north east,
        legend style={nodes={scale=0.8, transform shape}},
        legend image post style={scale=0.8},
        ]
    
        \addplot[
            fill=orange!50,
            draw=black,
            point meta=y,
            every node near coord/.style={inner ysep=5pt},
        ] 
        table [y] {
            x y
            1 0.65
            2 0.615
            3 0.647
            4 0.542
            5 0.582
            6 0.586
            7 0.484
            8 0.480
            9 0.480
        };

        \addplot[
            fill=violet!50,
            draw=black,
            point meta=y,
            every node near coord/.style={inner ysep=5pt},
        ] 
        table [y] {
            x y
            1 0.30
            2 0.395
            3 0.297
            4 0.27
            5 0.237
            6 0.199
            7 0.216
            8 0.167
            9 0.132
        };

    \draw ({rel axis cs:0,0}|-{axis cs:0,0}) -- ({rel axis cs:1,0}|-{axis cs:0,0});
    \legend{MAC*, DAFT MAC}
    \end{axis}
    \end{tikzpicture}
    \caption{
        Mean growth of TLT that starts from 2-step. 
        Bars denote arithmetic mean value of given interval.
        For example, the bar at 12 represents $\frac{TLT_{12} - TLT_{2}}{10}$. 
        The figure is linked with \cref{fig:step_shifts}.
    }
    \label{fig:growth_tlt}
\end{figure}
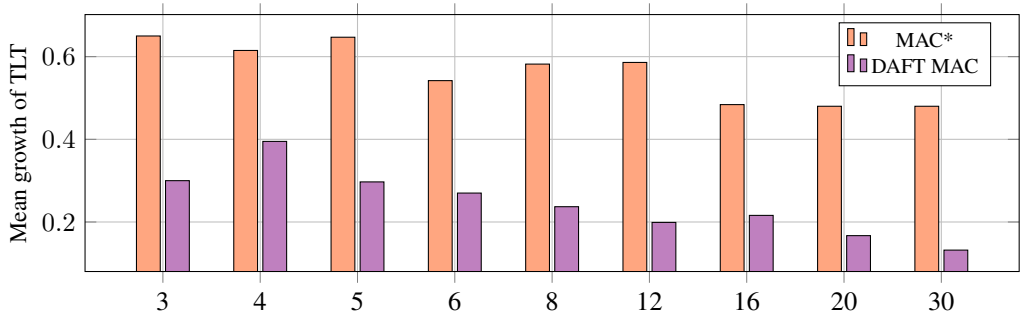

\begin{figure}[ht]
\centering
\input{contents/appendix_figures/fig_mac_diffseed.tex}
\caption{
    Attention maps from the other four 12-step MACs initialized with different seeds, 
    distributed over question \textit{"Are there more green blocks than shiny cubes?"}. 
    All of them perform similiarly to the model used in \cref{fig:interpretation} in terms of CLEVR validation accuracy. 
}
\label{fig:stability_mac}
\end{figure}

\begin{figure}[ht]
\centering
\input{contents/appendix_figures/fig_daftmac_diffseed.tex}
\caption{
    Attention maps from the other four 12-step DAFT MACs initialized with different seeds, 
    distributed over question \textit{"Are there more green blocks than shiny cubes?"}. 
    All of them perform similiarly to the model used in \cref{fig:interpretation} in terms of CLEVR validation accuracy. 
}
\label{fig:stability_daftmac}
\end{figure}

\begin{figure}[ht]
\centering
\input{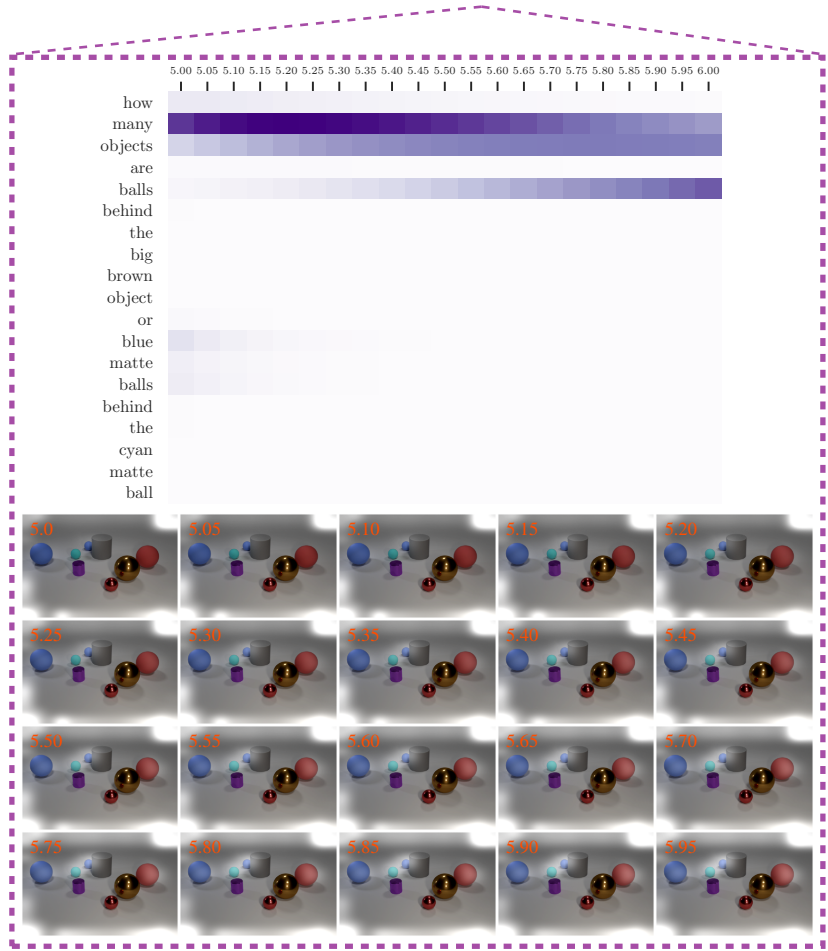}
\caption{
    Interpolation in-between steps. 
    Since the solution of IVP is a continuous function of time, 
    we can get a attention map for any given intermediate time value. 
    This fact enables infinitely fine-grained interpolation. 
    Also note that this is not a linear interpolation, see how the attention on \textit{many} reaches a maximum around 5.2 instead of on either end.
    }
\label{fig:interpolation}
\end{figure}

\begin{figure}[ht]
\centering
\input{contents/appendix_figures/fig_s8_compare_mac.tex}
\caption{
    Attention maps of 8-step MAC, distributed over question \textit{"How many objects are balls behind the big brown object or blue matte balls behind the cyan matte ball?"}. 
    This model achieves 99\% CLEVR validation accuracy.
}
\label{fig:compare_s8_mac}
\end{figure}

\begin{figure}[ht]
\centering
\input{contents/appendix_figures/fig_s8_compare_daftmac.tex}
\caption{
    Attention maps of 8-step DAFT MAC, distributed over question \textit{"How many objects are balls behind the big brown object or blue matte balls behind the cyan matte ball?"}. 
    This model achieves 99\% CLEVR validation accuracy.
}
\label{fig:compare_s8_daftmac}
\end{figure}

\begin{figure}[ht]
\centering
\input{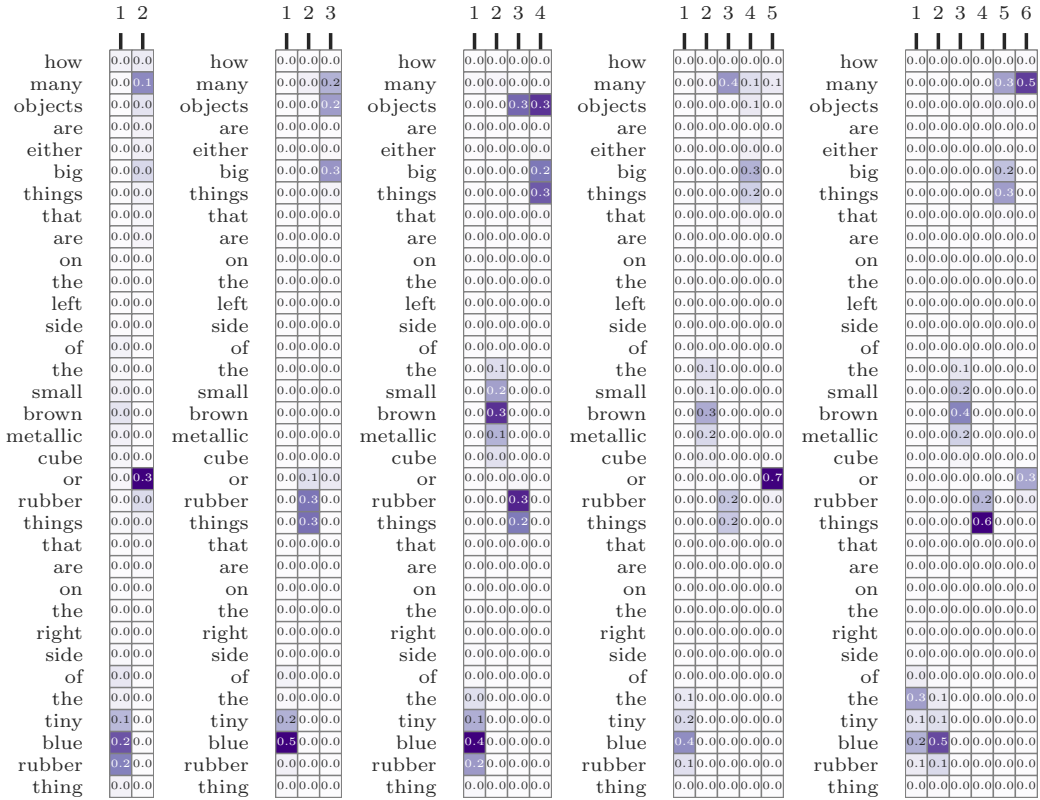}
\caption{
    Attention maps of DAFT MAC with $2$ to $6$ steps, distributed over the very long question \textit{"How many objects are either big things that are on the left side of the small brown metallic cube or rubber things that are on the right side of the tiny blue rubber thing?"}.
    Note that these five models are seperately initialized and thus have totally \textit{different} parameters. 
    The order of transition is unchanged among these completely separate models with different expressive power.
}
\label{fig:long_question}
\end{figure}

\begin{figure}[ht]
\centering
\input{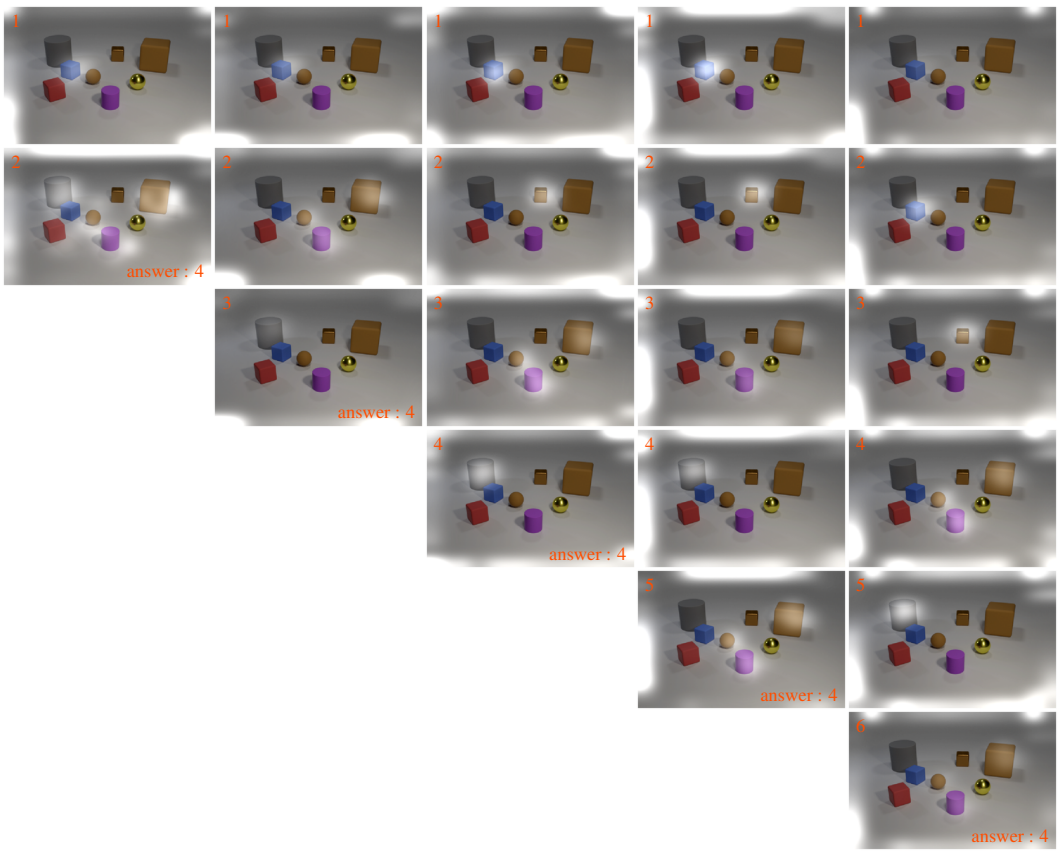}
\caption{
    Accompanying image attention maps for \cref{fig:long_question}.
}
\label{fig:long_question_image}
\end{figure}
\end{document}